\pdfoutput=1

\documentclass[11pt]{article}

\usepackage[preprint]{acl}

\usepackage{times}
\usepackage{latexsym}

\usepackage[T1]{fontenc}

\usepackage[utf8]{inputenc}

\usepackage{microtype}

\usepackage{inconsolata}

\usepackage{graphicx}

\usepackage{amsmath}
\usepackage{graphicx}
\usepackage{algorithm}
\usepackage[noend]{algpseudocode}
\usepackage{amsmath}
\usepackage[T1]{fontenc}
\usepackage{graphicx}
\usepackage{textcomp}
\usepackage{tikz}
\usetikzlibrary{shapes,arrows}
\usepackage{hyperref}
\usepackage{textcomp}
\usepackage{color}
\usepackage[utf8]{inputenc}
\usepackage{fancyvrb}
\usepackage{xcolor}
\usepackage{colortbl}
\usepackage{hhline}
\usepackage{lscape}
\usepackage{upquote}
\usepackage{array}
\usepackage{adjustbox}
\usepackage{rotating}
\usepackage{stmaryrd}
\usepackage{pifont}
\usepackage{multirow}
\usepackage{wrapfig}
\usepackage{verbatim}
\usepackage{colortbl}
\usepackage{rotating}
\usepackage{lipsum} 
\usepackage{tabularx}
\usepackage{booktabs} 
\usepackage{threeparttable}
\usepackage{lineno,hyperref}
\usepackage{colortbl}
\usepackage{rotating}
\usepackage{pgfplots,pgfplotstable}
\usepackage{scalefnt}

\newcolumntype{L}[1]{>{\raggedright\arraybackslash}m{#1}}
\newcolumntype{C}[1]{>{\centering\arraybackslash}m{#1}}
\newcolumntype{R}[1]{>{\raggedleft\arraybackslash}m{#1}}

\fvset{commandchars=\\\{\}}

\definecolor{Green}{rgb}{0.0, 0.56, 0.0}
\definecolor{redtext}{RGB}{255,0,0}
\definecolor{Gray}{gray}{0.85}

\newcommand{\CodeIn}[1]{\begin{small}\texttt{#1}\end{small}}
\newcommand{\NoType}[1]{} 
\newcommand{\Space}[1]{}

\newcommand{\DefMacro}[2]{\expandafter\newcommand\csname rmk-#1\endcsname{#2}}
\newcommand{\UseMacro}[1]{\csname rmk-#1\endcsname}

\author{Anahita Samadi \and Allison Sullivan \\
         The University of Texas at Arlingotn \\ 
         \texttt{anahita.samadi@mavs.uta.edu} \\
         \texttt{and allison.sullivan@uta.edu}
         }

\title{Evaluating Text Classification Robustness to Part-of-Speech Adversarial Examples}

\begin{document}

\maketitle

\begin{abstract}
As machine learning systems become more widely used, especially for safety critical applications, there is a growing need to ensure that these systems behave as intended, even in the face of adversarial examples. Adversarial examples are inputs that are designed to trick the decision making process, and are intended to be imperceptible to humans. However, for text-based classification systems, changes to the input, a string of text, are always perceptible. Therefore, text-based adversarial examples instead focus on trying to preserve semantics. Unfortunately, recent work has shown this goal is often not met. To improve the quality of text-based adversarial examples, we need to know what elements of the input text are worth focusing on. To address this, in this paper, we explore what parts of speech have the highest impact of text-based classifiers. Our experiments highlight a distinct bias in CNN algorithms against certain parts of speech tokens within review datasets. This finding underscores a critical vulnerability in the linguistic processing capabilities of CNNs. 
\end{abstract}

\section{Introduction}\label{sec:intro}

Machine learning systems are increasing in popularity. As a result, these systems are increasingly applied to help address both everyday and safety critical tasks. Therefore, we need to ensure that these systems are robust. One common application of machine learning systems is for text classification, which is used for tasks like sentiment analysis and spam detection \cite{arif2018sentiment}. Unfortunately, recent work has shown that text classification systems are susceptible to adversarial attacks~\cite{papernot2016crafting,jia2017adversarial,belinkov2017synthetic,glockner2018breaking,iyyer2018adversarial}.


Adversarial examples are specifically crafted inputs that are designed to exploit vulnerabilities in the classifier's decision-making process, with the aim of leading to incorrect predictions. These examples can be generated by introducing imperceptible perturbations to the input data, causing the classifier to misclassify or make erroneous judgments. Adversarial examples were first explored for computer vision, which demonstrated that two images that looked exactly the same to a human can produce different predictions form a classifier~\cite{szegedy2013intriguing}. Since then, adversarial examples have been applied to other fields such as image and speech recognition~\cite{zhang2020adversarial,kurakin2018adversarial,papernot2017practical}.

However, text based adversarial examples are identifiable to humans, not imperceptible, as two different sequences of text are never entirely indistinguishable. Therefore, researchers instead focus on designing attacks that look to preserve the semantic meaning of the text, such as replacing works with synonyms~\cite{alzantot2018generating,jin2020bert,kuleshov2018adversarial,papernot2016crafting,ren2019generating}. However, recent work has shown that even these attacks rarely succeed at preserving semantics~\cite{hauser2021bert,morris2020reevaluating}.



Given that semantics is easily lost when making transformations to text, we need to better understand what aspects of a text are likely to influence the classification system in order to have a more narrow subset of the text to focus on to figure out how to reliably generate meaningful, non-trivial adversarial examples. Therefore, in this paper, we focus on examining the impact of different parts of speech (POS) on the decision-making process of Convolutional Neural Network (CNN) classifiers. 
Accordingly, in this paper, we make the following contributions:

\newenvironment{Contributions}{}{} 
\newcommand{\Contribution}[1]{\textbf{#1:}} 

\begin{Contributions}

\Contribution{Dataset for Adversarial Algorithm Training} We create a dataset to evaluate the performance of text-based machine learning classifiers.

\Contribution{Adversarial Neural Network} 
We introduce a new adversarial neural network. This network is trained to learn the vulnerabilities and decision-making patterns of the underlying classifier.

\Contribution{Generating Adversarial Examples} Utilizing our adversarial neural network, we generate adversarial examples by manipulating the input text based on the parts of speech present.

\Contribution{Evaluation} We explore the impact of removing 1\%, 5\%, 10\%, and 15\% of words from targeted parts of speech on a CNN model's predictive reliability across three commonly used datasets (Amazon, Yelp, and IMDB). The results highlight that a CNN classifiers have a bias towrds verbs, nouns and adjectives, and finding a small subset of these to manipulate is likely to consistently produce effective adversarial examples. 

\end{Contributions}

\section{Background}\label{sec:bg}

In this section, we describe key concepts of text classification and text sentiment analysis.

\textbf{Text classification.}
Text classification, the task of categorizing text into classes, is a critical function in natural language processing with various applications such as sentiment analysis, spam detection, language detection, document classification and more \cite{arif2018sentiment}. This process has traditionally been performed using statistical methods such as Naive Bayes, Support Vector Machines, and Decision Trees, or reinforcement learning-based classifiers, including Boosting and Bagging \cite{kowsari2019text,arif2018sentiment}. However, the advent of deep learning has introduced models like Convolutional Neural Networks, Recurrent Neural Networks (RNNs), Long Short-Term Memory (LSTM), and Transformers which have shown remarkable accuracy in capturing complex patterns in text data \cite{oshea2015introduction,SHERSTINSKY2020132306,DBLP:journals/corr/VaswaniSPUJGKP17}. In text analysis, techniques such as Bag-of-Words (BoW), Term Frequency-Inverse Document Frequency (TF-IDF), and word embeddings like Word2Vec and GloVe have been developed to represent data as understandable input for machine learning models.

\textbf{CNN for Text Sentiment Analysis.}
Text sentiment analysis is a specific text classification task in which a system aims to classify a text as having positive, negative, or neutral emotion. CNNs are a class of deep feed-forward neural networks that require minimal preprocessing, due to the use of multi-layer perceptrons. CNNs treat text as a structured sequence of tokens and apply convolutions to capture local features. \citeauthor{kim2014convolutional} demonstrates that CNNs can be applied to text classification tasks. Since then, there  have been several frameworks developed which utilize CNNs to perform various text sentiment analysis tasks~\cite{zhang2015character,zhang2015sensitivity,zhang2015text,liu2019interpretable,ce2020interpretability}.

We employee a CNN-based text classification model that uses GloVe pre-trained embeddings in the initial embedding layer to map input sequences into dense vector representations. A subsequent 1D convolutional layer captures local textual features, which is followed by batch normalization and dropout (50\%) to enhance training stability and mitigate overfitting. A global max-pooling layer reduces the dimensionality by retaining the most salient features, which are then fed into a dense layer with a sigmoid activation function for binary classification. The model is optimized using Adam \cite{kingma2015adam} and binary cross-entropy loss:

\noindent\resizebox{\columnwidth}{!}{
$L(y, \hat{y}) = -\frac{1}{N} \sum_{i=1}^N \left[ y_i \log(\hat{y}_i) + (1 - y_i) \log(1 - \hat{y}_i) \right]$
}

Early stopping is incorporated to terminate training when validation loss improvement ceases, ensuring the best model weights are retained.

\begin{figure*}[t]
 \centering
  \includegraphics[width=11cm]{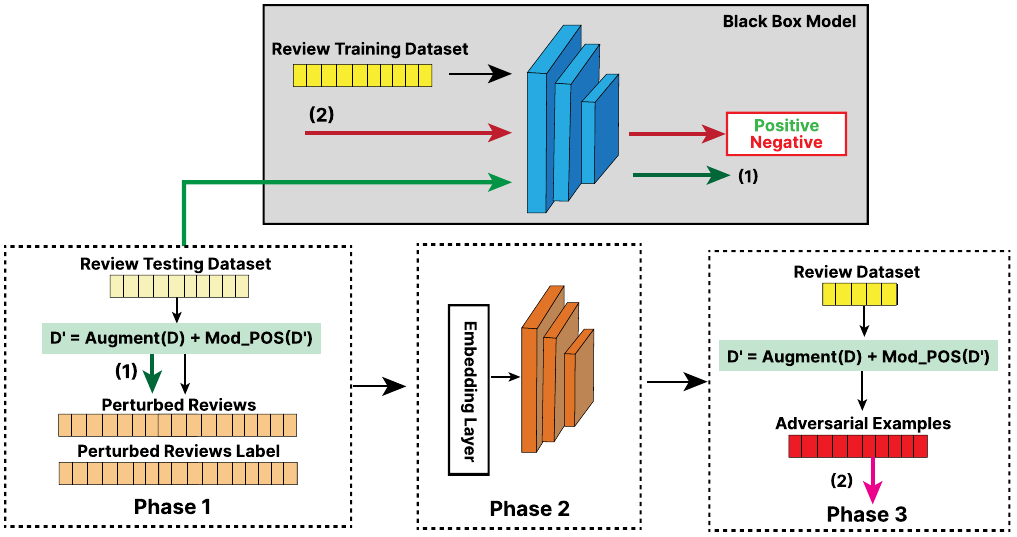}
  \caption{Overview of 3 Phase Adversarial Example Generation Technique}
  \label{fig:pipeline}
\end{figure*}
\section{Threat Model}\label{sec:threat}


In this paper, we outline a method to systematically generate adversarial examples with the objective of evaluating the robustness of CNNs employed in the classification of text-based reviews. Our methodology is predicated on a black-box attack paradigm, where the adversary does not possess access to the internal parameters or gradients of the target model. Instead, the adversary operates under the assumption that they can only observe and analyze the model’s output given specific inputs.

The threat model is designed to reflect realistic adversarial conditions, wherein the adversary is constrained by limited access to labeled textual data. This constraint is reflective of practical scenarios, where the acquisition of a comprehensive, labeled corpus of sensitive information is often challenging and resource-intensive. The adversary's objective within this model is to identify and exploit vulnerabilities in the CNN by inducing misclassifications through minimal modifications to the input text.
 The design of this threat model underscores its applicability in evaluating the robustness of text classification systems under conditions that mirror realistic adversarial scenarios.






\section{Technique}\label{sec:tech}
Our primary aim is to illustrate how certain parts of speech can affect the accuracy of CNN classifers, leading to misleading results. To achieve this, we present an adversarial example generation framework which consists of three phases, an overview of which can bee seen in Figure~\ref{fig:pipeline}. In Phase 1, we explore the current effectiveness of CNN classifiers and the impact of different parts of speech in order to gather relevant information to inform the design choices for our adversarial attack. In Phase 2, we design an Adversarial Neural Network that will delete select keywords from the targeted part(s) of speech. In Phase 3, we use the model created in Phase 2 to generate adversarial examples. 


\subsection{Phase 1: Creating the dataset to train the adversarial algorithm}

To explore which parts of speech will effect the classification accuracy of a neural network, we propose a new approach to generate an adversarial dataset from a primary dataset in order to feed into our adversarial neural network.  Algorithm~\ref{phase1} shows the details of phase 1 of our approach, which consists of the following steps:

\textbf{Step 1:} We use the CNN classifier outlined in Section~\ref{sec:bg} to 
calculate the model's initial accuracy for our datasets under consideration (Lines 14-15). This process enables us to establish a baseline for each dataset, which gives us a measurable starting point to help determine the impact of different adversarial attacks.

%
%
        
\textbf {Step 2:} To prevent data leakage, we randomly collect reviews from the test portion of the dataset for analysis to create our adversarial training dataset (Lines 7-9). This ensures that the data used in the adversarial approach is distinct from the training data used in the initial model, thereby minimizing the potential for bias and overfitting.


\begin{algorithm}[t]
    \caption{ Data Set Creation for the Adversarial Example Algorithm}\label{phase1}
   \small
   \begin{flushleft}  \textbf{Input: } Review data set \end{flushleft} 
    \begin{flushleft}\textbf{Output:} Sample reviews without random tokens, New labels to guide adversarial algorithm  \end{flushleft}
\scriptsize
   \begin{algorithmic}[1]

    \State $PredictedLabels \gets TestSetClassificationResults$ 
    \State $OriginalLabels \gets TestSetLabels$ 
    \State $Ratio \gets WordsDeletionPercentage$ 
    \State Replicate and shuffle each review in test set and corresponding labels in $OriginalLabels$ and $Predictedlabels$ 
\\
    \Procedure{ Create\_Adversarial\_XTrain}{$ReplicatedXtrain$}
        \For {each $Review \in ReplicatedXtrain$}
            \State Collect $Ratio$ Random Token in VERB, NOUN, ADJ  category from ${Review}$
            \State $DelToken \gets RandomTokens$
            \State $AdvReview \gets ManipulatedReview$

        \EndFor
    \State \textbf{return}\ $AdvReview$, $DelToken$
    \EndProcedure
\\    
    \Procedure{Create\_Adversarial\_Labels }{$PredictedLabels$,$OriginalLabels$,$AdvReview$}
        \State  classification Prediction with $AdvReview$
        \State $AdvPred \gets AdvReviewClassificationResults$ 
        \State $AndLabels\gets PredictedLabels \land  OriginalLabels$
        
        \For {each ${rec}\in PredictedLabels$}
            \If{$AdvPred\neq AndLabels$} 
                \State $AdvLabel \gets 1$
            \Else
                \State $AdvLabel \gets 0$                
            \EndIf
        \EndFor
 
    \State \textbf{return}$AdvLabel$
    \EndProcedure 
    \end{algorithmic}
    \end{algorithm}

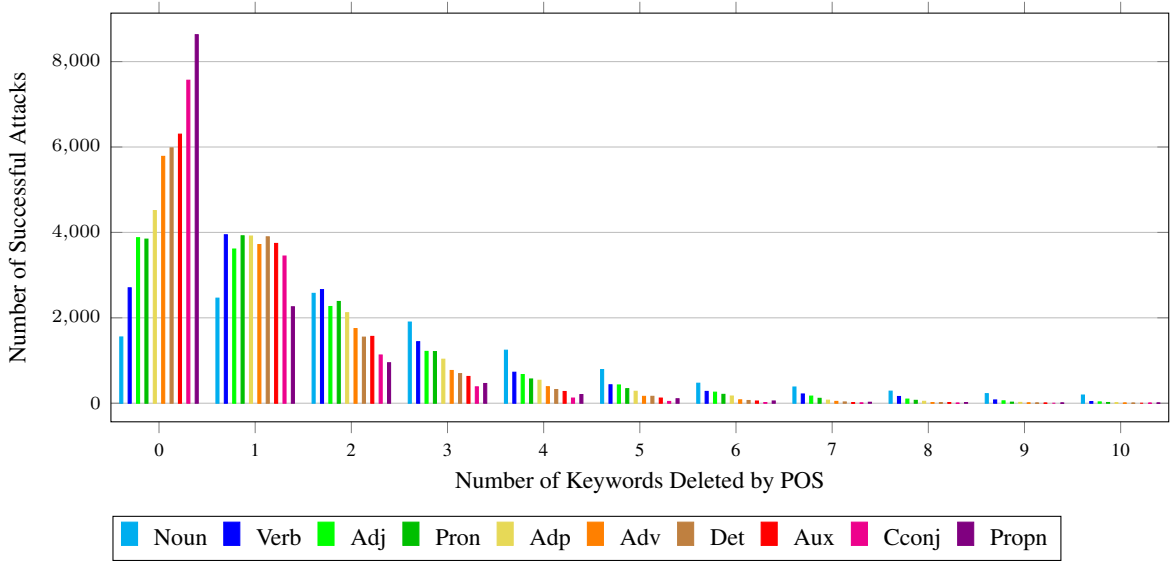
\begin{figure*}[t]
 \centering
  \begin{tikzpicture}
\begin{axis}
[
    ybar,
     width=15.5cm, bar width=0.04cm,
    height=7cm,
    enlargelimits=0.05,
    ymin=0, ymax=8700, 
    legend style={at={(0.45,-0.23)},
      anchor=north,column sep=0.8ex,legend columns=-1},
    yticklabel style={font=\scriptsize}, xticklabel style={font=\scriptsize},
    ylabel={\small Number of Successful Attacks}, xlabel={\small Number of Keywords Deleted by POS},
    xtick=data,
    symbolic x coords={0,1,2,3,4,5,6,7,8,9,10},
    ymajorgrids= true
]
    
\addplot[style={cyan, fill=cyan}, single ybar legend] 
    coordinates {(0,1552) (1,2461) (2,2573) (3,1901) (4,1242) (5,788) (6,469) (7,380) (8,283) (9,224) (10,192)};
\addplot[style={blue, fill=blue}, single ybar legend] 
    coordinates {(0,2703) (1,3945) (2,2660) (3,1441) (4,726) (5,432) (6,277) (7,217) (8,158) (9,77) (10,39)};
\addplot[style={green, fill=green}, single ybar legend]
    coordinates {(0,3876) (1,3611) (2,2265) (3,1214) (4,672) (5,428) (6,259) (7,167) (8,97) (9,56) (10,31)};
\addplot[style={green!75!black, fill=green!75!black}, single ybar legend]
    coordinates {(0,3843) (1,3921) (2,2383) (3,1209) (4,568) (5,341) (6,208) (7,115) (8,68) (9,25) (10,17)};
\addplot[style={yellow!75!gray, fill=yellow!75!gray}, single ybar legend]
    coordinates {(0,4511) (1,3915) (2,2120) (3,1031) (4,538) (5,280) (6,169) (7,72) (8,45) (9,19) (10,9)};
\addplot[style={orange, fill=orange}, single ybar legend]
    coordinates {(0,5781) (1,3716) (2,1748) (3,767) (4,387) (5,161) (6,80) (7,43) (8,16) (9,6) (10,3)};
\addplot[style={brown, fill=brown}, single ybar legend]
    coordinates {(0,5980) (1,3897) (2,1548) (3,693) (4,321) (5,162) (6,64) (7,30) (8,11) (9,4) (10,1)};
\addplot[style={red, fill=red}, single ybar legend]
    coordinates {(0,6301) (1,3742) (2,1565) (3,627) (4,275) (5,121) (6,52) (7,12) (8,14) (9,2) (10,0)};
\addplot[style={magenta, fill=magenta}, single ybar legend]
    coordinates {(0,7564) (1,3447) (2,1130) (3,384) (4,121) (5,42) (6,16) (7,5) (8,1) (9,0) (10,1)};
\addplot[style={violet, fill=violet}, single ybar legend]
    coordinates {(0,8630) (1,2259) (2,948) (3,462) (4,203) (5,105) (6,53) (7,23) (8,16) (9,5) (10,4)};
    
\legend{\small Noun, \small Verb, \small Adj, \small Pron, \small Adp, \small Adv, \small Det, \small Aux, \small Cconj, \small Propn};    
\end{axis}
\end{tikzpicture}
  \caption{Number of successful random attacks with specific number of keywords present.}
  \label{fig:recalgo2}
\end{figure*}

\textbf{Step 3:} To help narrow our focus for adversarial example generation, we performed a series of random attacks against the CNN classifier to investigate the impact of 10 different parts of speech on the decision-making process. From 990 reviews from the IMDB dataset, we replicated each review 100 times, creating 99,000 reviews from which we gathered a set of ten random words to delete. We then flagged whether the random attack was successful, which was the case for 12,873 of these reviews. For successful attacks, we recorded which parts of speech were deleted, the results of which are displayed in Figure~\ref{fig:recalgo2}. 

Our findings reveal that among the successful random attacks against the classifier, the classifier was most vulnerable to the removal of nouns and verbs followed by pronouns and adjectives, with all other parts of speech having notably less of an impact. In particular, having zero nouns, verbs, adjectives, and pronouns deleted is associated with the four lowest numbers of successful random attacks (0 keywords). On the other hand, deleting at least 1 noun or verb contributed to 81.7\% and 77.5\% respectively of the effective random attacks. Adjectives and pronouns were nearly as equally effective as one another: deleting at least 1 adjective or pronoun contributed to 68.4\% and 68.8\% respectively of the effective random attacks.

Given the similarities of use cases between nouns and pronouns, we decided to focus on nouns, verbs and adjectives in order to devise more effective attacks.  


\textbf{Step 4:} Due to variations in the length of individual reviews within the target datasets, we sought to ensure the fairness of our experiments by examining the impact of word deletion ratios within the noun, verb, and adjective categories. Specifically, we explore  deletion ratios of 1\%, 5\%, 10\%, and 15\% of the total number of nouns, verbs, and adjectives in each review (Lines 1-11). 

To ensure comprehensive coverage of diverse data, we examine 100 different combinations of word deletions for each ratio across all reviews in the datasets to achieve adequate diversity in permutations. For each review, we identified all tokens (VERB, NOUN, ADJ). For a given ratio \( r\% \) (e.g., \( r = 1\% \)), we calculated the number of tokens to delete as \( n = \left\lceil \frac{r}{100} \times \text{length of review} \right\rceil \). We then randomly selected \( n \) tokens from the identified set of tokens (Lines 7-8).  

To mitigate the loss of information and potential disruption to the text structure, we replace the collected random words with a placeholders token. We also store the deleted words and manipulated reviews in \CodeIn{DelToken} and \CodeIn{AdvReview} respectively for future use (Lines 9-10).
        

    
\textbf {Step 5:} Then, we feed the CNN model that was trained in \textbf{Step 1} with the adversarial dataset that we generated in \textbf{Step 4}. We store the prediction values the CNN model makes over the adversarial dataset in \CodeIn{AdvPred}. (Lines 14-15)

\textbf {Step 6:} We generate labels for our proposed adversarial approach (Lines 17-21) . We assign a value of 1 to indicate that an adversarial example successfully fools the classifier, and we assign a value of 0 if adversarial example does not have any impact on the classifier's output. 

To evaluate the performance of the classifier using adversarial examples, it is important to ensure that the adversarial dataset is appropriately labeled. Our schema for generating labels takes into account two important considerations. Firstly, the labels assigned to the adversarial dataset should not improve the classifier's prediction, which is also known as a false positive. Secondly, the labels assigned to the adversarial dataset should have a different value from the classifier's prediction. In other words, the value of adversarial labels should contrast with the classifier's prediction value only if the classifier prediction was the same as the original corresponding review label.
By following our schema, we ensure that the adversarial examples are designed to mislead the classifier, and do not help improve the classifiers' performance by predicting wrong values and reversing them. 


\textbf {Step 7:} We organize the adversarial review dataset and the corresponding generated labels in accordance with the restrictions specified in \textbf{Step 6} (Lines 22). The organized dataset is then used as the input to Phase 2 of our framework.

\subsection{Phase 2: Adversarial Neural Network.} 

\begin{figure}[t]
 \centering
    \includegraphics[width=0.80\columnwidth]{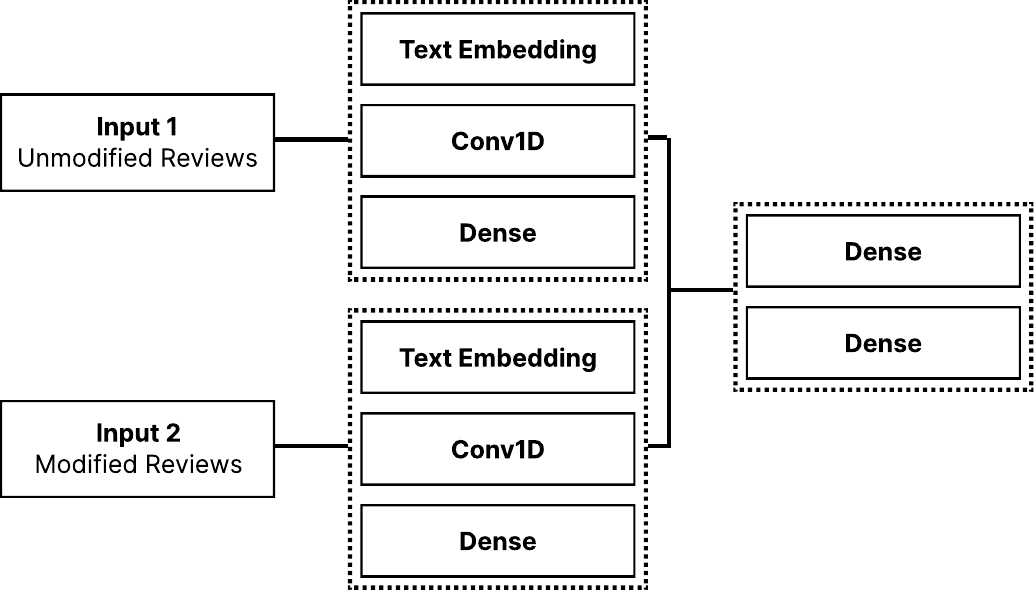}
    \caption{Phase 2 Adversarial Algorithm Architecture}
    \label{fig:recalgo}
\end{figure}


In this phase, we investigate the optimal words to remove from the review dataset to create an adversarial dataset. To accomplish this, we utilize a Neural Network to train an adversarial model, an overview of which is depicted in Figure~\ref{fig:recalgo}. The original reviews are compared to the same reviews with selected keywords removed. These two inputs are fed into separate branches of the network. The output of each branch is then concatenated into a single input for the last layers of the network. 

By doing so, we mathematically proved the relationship between deleted words and the overall context of the review. The output of this network is a binary label. A label of 1 indicates that the manipulated review can successfully deceive the classifier algorithm and that the deleted words were crucial to the decision. In contrast, a label of 0 indicates that the chosen words to delete were not effective in deceiving the classifier. The resulting adversarial neural network model is saved for Phase 3, where the model will be used to create an adversarial examples.

\subsection{Phase 3: Generating adversarial examples} 
In this phase, we present an approach to generate adversarial examples using the model developed in Phase 2, where the attacker learned the classifier's vulnerabilities and decision-making patterns. Specifically, in this step we modify randomly selected reviews based on the pattern detected in the previous step. 

\begin{algorithm}[t]
    \caption{Algorithm to Generate Adversarial Examples 
   }\label{phase3}
    \small
   \begin{flushleft}  \textbf{Input: } Random Reviews \end{flushleft} 
    \begin{flushleft}\textbf{Output:} Adversarial Examples  \end{flushleft}
 \scriptsize
    \begin{algorithmic}[1]
    \State $TrainAdvReviews \gets AdversarialReviews$ 
    \State $Ratio \gets WordsDeletionPercentage$ 
    \State Replicate and shuffle each review in $AdvReviews$  and corresponding label in $TrainAdvReviews$ 100 Times
\\
    \Procedure{ Create\_Manipulated\_Review}{$ReplicatedADVtrain$}
        \For {each $Review \in ReplicatedADVtrain$}
            \State Collect $Ratio$ Random Token in VERB,NOUN,ADJ category from ${Review}$
            \State $DelToken \gets RandomTokens$
            \State $AdvReview \gets ManipulatedReview$

        \EndFor
    \State \textbf{return}$AdvReview$, $DelToken$
    \EndProcedure
\\    
    \Procedure{ Create\_Adversarial\_Example}{$AdvReview$}
        \State  use the NN model in phase 2 to predict $AdvReview$
        \State $AdvPred \gets AdvReviewClassificationResults$ 
        \State Sort reviews based on corresponding prediction value in descending order    
        \State Collection of top 100 adversarial examples  
        \State $AdvExpReviews \gets CollectedAdvReview$ 
 
    \State \textbf{return}$AdvExpReviews$  
    \EndProcedure

    \end{algorithmic}
    \end{algorithm}

Algorithm~\ref{phase3} shows the details of Phase 3 of our approach, which consists of the following steps:
        
\textbf {Step 1:} We gather 100 random reviews from the target dataset. Then, we duplicate each review a hundred times, resulting in a set of replicated reviews stored in the variable \CodeIn{ReplicatedADVtrain} (Line 3).

\begin{figure*}[t]
 \centering
    \begin{tikzpicture}
\begin{axis}[
    ybar,
    width=15cm, bar width=0.1cm,
    height=5cm,
    enlargelimits=0.05,
    ymin=0, ymax=100, ytick={0,10,20,30,40,50,60,70,80,90,100},
    legend style={at={(0.5,-0.3)},
      anchor=north,column sep=0.8ex,legend columns=-1},
    yticklabel style={font=\scriptsize}, xticklabel style={font=\scriptsize},
    ylabel={\textbf{Test Accuracy (Adversarial)}}, xlabel={\textbf{Combinations of Parts-of-Speech}},
    xtick=data,
    symbolic x coords={VERB,ADJ,NOUN,VERB\_ADJ,VERB\_NOUN,ADJ\_NOUN,VERB\_ADJ\_NOUN},
    ymajorgrids= true
    ]
    
\addplot[draw=blue,thick, smooth, tension=0.001,mark=*,mark options={color=blue,scale=0.8}] 
    coordinates {(VERB,84) (ADJ,82) (NOUN,71) (VERB\_ADJ,79) (VERB\_NOUN,75) (ADJ\_NOUN,86) (VERB\_ADJ\_NOUN,54)};

\addplot[draw=Green,thick, smooth, tension=0.001,mark=*,mark options={color=Green,scale=0.8}] 
    coordinates {(VERB,82) (ADJ,49) (NOUN,80) (VERB\_ADJ,73) (VERB\_NOUN,76) (ADJ\_NOUN,73) (VERB\_ADJ\_NOUN,73)};

\addplot[draw=orange,thick, smooth, tension=0.001,mark=*,mark options={color=orange,scale=0.8}] 
    coordinates {(VERB,54) (ADJ,88) (NOUN,84) (VERB\_ADJ,77) (VERB\_NOUN,85) (ADJ\_NOUN,78) (VERB\_ADJ\_NOUN,62)};

\addplot[draw=purple,thick, smooth, tension=0.001,mark=*,mark options={color=purple,scale=0.8}] 
    coordinates {(VERB,87) (ADJ,63) (NOUN,86) (VERB\_ADJ,67) (VERB\_NOUN,83) (ADJ\_NOUN,71) (VERB\_ADJ\_NOUN,61)};

\addplot[draw=red,thick, dashed,smooth, tension=0.001] 
    coordinates {(VERB,97) (ADJ,97) (NOUN,97) (VERB\_ADJ,97) (VERB\_NOUN,97) (ADJ\_NOUN,97) (VERB\_ADJ\_NOUN,97)};
\scriptsize
\legend{1\% Perturbation, 5\% Perturbation, 10\% Perturbation, 15\% Perturbation, Before Attack};    
\end{axis}
\end{tikzpicture}
    \caption{Attack Against the IMDB Dataset for Different POS}
    \label{fig:Pos}
\end{figure*}
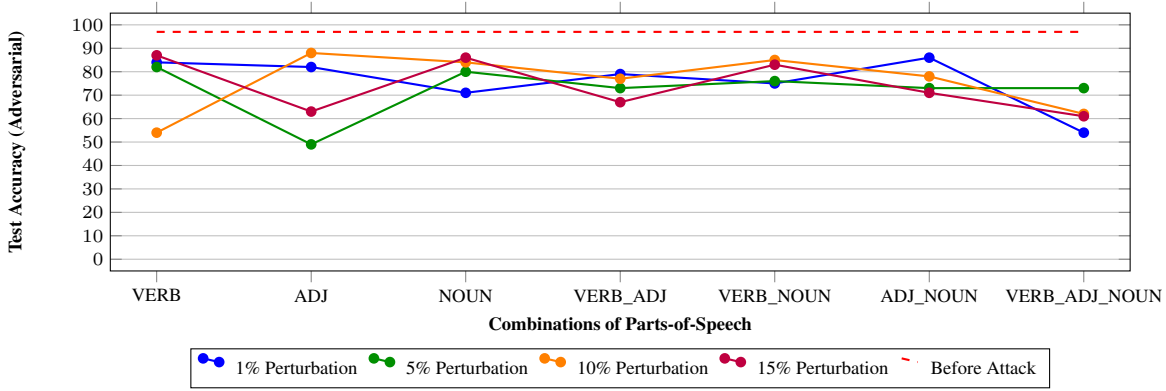

\textbf {Step 2:} As with Phase 1, we next collect random words consisting of nouns, verbs, and adjectives as candidates for removal. We then conduct a series of experiments using deletion ratios of 1\%, 5\%, 10\%, and 15\%  of the total count of nouns, verbs, and adjectives in each review across all datasets. When deleting words, our experiments explore all the different combinations of adjectives, nouns and verbs. These configurations range from deleting from just one part of speech to deleting from different paired parts of speech to deleting form all three parts of speech.
Just as in \textbf{Phase 1}, we store the removed words and the altered replicated reviews in \CodeIn{DelToken} and \CodeIn{AdvReview}, respectively (Lines 5-10).  

\textbf {Step 3:} Utilizing the model trained in Phase 2, we predict the outcomes for the \CodeIn{ReplicatedADVtrain} dataset, which was generated by \textbf{Step 1} of Phase 3 (Lines 13-14).

\textbf {Step 4:} We arrange the reviews in descending order based on their respective prediction values (Line 15).

\textbf {Step 5:} We compile the top 100 reviews exhibiting the highest prediction values and store them as adversarial examples. These adversarial examples are then used to attack the primary CNN model that we evaluated in \textbf{Step 1} of Phase 1 (Lines 16-18). 

\section{Evaluation}\label{sec:eval}
In this section we address the following research questions.

\begin{itemize} \itemsep=0em

    \item \textbf{RQ1:} What is the impact of deleting different parts of speech?
    
    \item \textbf{RQ2:} How does the efficacy of deleting different parts of speech vary over different datasts?

    \end{itemize}

\subsection{Set Up}
To evaluate the efficacy of our adversarial examples, we utilize three well-known datasets of reviews, which are commonly used to evaluated text sentiment analysis techniques.

First, as our primary dataset, we use the IMDB dataset, which contains user's reviews of different movies and media. The dataset is intended for binary sentiment analysis. The dataset is balanced, with an equal distribution of positive and negative reviews. Each segment, training and testing, comprises of 10,000 reviews~\cite{kaggleimdb}. 

Second, we use the The Amazon Polarity dataset. The dataset is intended for binary sentiment classification and contains 2 million reviews split evenly between positive and negative sentiments. The Amazon Reviews dataset spans 18 years, includes around 35 million reviews up to March 2013, and features product details, user information, ratings, and plaintext reviews~\cite{kagglend}. 

Lastly, we use the Yelp Polarity dataset, which organizes multi-star rating reviews to be applicable to binary sentiment analysis. 
The dataset categorizes Yelp reviews into two classes based on polarity: negative (Class 1) for reviews with 1 or 2 stars, and positive (Class 2) for reviews with 3 or 4 stars. It includes 560,000 training samples (280,000 per class) and 38,000 testing samples (19,000 per class)~\cite{kaggleyelp}. 



\begin{table*}[t]
\caption{In this table, we present the outcome of applying our approach to a randomly selected review from the IMDB dataset. Our algorithm suggests the deletion of words highlighted in red. Upon their removal, we demonstrate the ability to mislead the predictive accuracy of CNN classification algorithm.}
\label{textdeltable2}
\begin{threeparttable}
\centering
\small 
\setlength{\tabcolsep}{2pt} 
\resizebox{1.9\columnwidth}{!}{\begin{tabularx}{\textwidth}{p{0.09\textwidth} p{0.74\textwidth} p{0.15\textwidth}} 
\toprule
\textbf{IMDB} & \textbf{review deleting ratio \%1 (red = deleted tokens)} & \textbf{Model Prediction} \\
\midrule
ADJ \%1 &  my complaints here concern the movie pacing and the material at hand while using archival film and letters lends the film fresh and interesting perspective too often the material selected to highlight simply isn very interesting \textcolor{redtext}{such} as when goebbels complains about this or that ailment tc or the ad nauseam footage of his small german hometown also the movie crawls along in covering and then steams through the war years in sum the film is little better than history channel documentary with the exception that the filmmaker has slightly greater sensibility than your average history channel documentary editor and thus can more artfully arrange the details of goebbels life still found it wanting & Negative $\rightarrow$ Positive \\
NOUN \%1 & my complaints here concern the movie pacing and the material at hand while using archival film and letters lends the film fresh and interesting perspective too often the material selected to highlight simply isn very interesting such as when goebbels complains about this or that ailment tc or the ad nauseam footage of his small german hometown also the movie crawls along in covering and then steams through the war years in sum the film is little better than history \textcolor{redtext}{channel} documentary with the exception that the filmmaker has slightly greater sensibility than your average history \textcolor{redtext}{channel} documentary editor and thus can more artfully arrange the \textcolor{redtext}{details} of goebbels \textcolor{redtext}{life} still found it wanting & Negative $\rightarrow$ Positive \\
VERB \%1 &  my complaints here concern the movie pacing and the material at hand while using archival film and letters lends the film fresh and interesting perspective too often the material selected to \textcolor{redtext}{highlight} simply isn very interesting such as when goebbels complains about this or that ailment tc or the ad nauseam footage of his small german hometown also the movie crawls along in covering and then steams through the war years in sum the film is little better than history channel documentary with the exception that the filmmaker has slightly greater sensibility than your average history channel documentary editor and thus can more artfully arrange the details of goebbels life still found it wanting & Negative $\rightarrow$ Positive \\
\bottomrule
\end{tabularx}}
\end{threeparttable}
\end{table*}

\subsection{RQ1: Deleting different parts of speech.}
To explore the impact of different parts of speech on the reliability of CNN models in interpreting and classifying text, we explore seven configurations: removing verbs-only, adjectives-only, nouns-only, verbs-adjectives, verbs-nouns, adjectives-nouns, and verbs-adjectives-nouns. For these experiments, we focused on the IMDB dataset and four levels of perturbation: 1\%, 5\%, 10\%, and 15\%. Figure~\ref{fig:Pos} displays our results, with the x-axis representing our configurations and the y-axis representing the test accuracy.  

Across all configurations, the verbs-adjectives-nouns combination is overall the most effective attack, decreasing the accuracy of the CNN classifier from 97\% to between 54\% and 73\% depending on the perturbation. Interestingly, the 1\% perturbation, which deletes the least amount of words, was the most effective, reducing the accuracy to 54\%, a 44.32\% reduction in accuracy compared to the baseline. Overall, increasing the number of keywords deleted did not have a consistently positive impact on the performance. For the noun-only, and verb-noun configurations, the success of the attack decreases with a higher perturbation. Meanwhile, for several other configurations, there is a ping-pong effect where the accuracy decreases, increases and then decreases again. This created a few outliers - such as verb-only 10\% and adjective-only 5\% - where these configurations produced a notably decrease in accuracy of the classifier compared to the surrounding perturbations.




To get further insight into this, Table~\ref{textdeltable2} showcases an effective attack using the 1\% adjective-only, 1\% noun-only, and 1\% verb-only configuration on a review from the IMDB dataset. In two of these three configuration, adjective-only and verb-only, the removal of just a single token is effective to misled the CNN's prediction. Meanwhile, for 1\% noun-only, the removal of just 4 token is effective. The noun-only configuration removes more words as there are more nouns overall in the example review than adjectives and verbs. Combined with our finding in Figure~\ref{fig:Pos}, our results highlight (1) the CNN classifier is not more vulnerable to more deleted words, but rather certain verb(s), noun(s) or adjective(s) can have greater impact on the perceived sentiment of the text, and (2) finding a critical combination of verbs, nouns and adjectives to manipulate is likely to consistently lead to high value adversarial examples. 

\subsection{RQ2: Performance across datasets.}

To explore how these results might generalize, we explore the performance of our adversarial example generation framework across two additional datasets Amazon and Yelp. For each dataset, we explore the best overall configuration (noun-verb-adjective) across our four levels of perturbation: 1\%, 5\%, 10\%, and 15\%. The impact on accuracy can be see in Table~\ref{tab:result}, which also includes results for the IMDB dataset from RQ1 for comparison. The highest reduction in accuracy over the associated baseline per perturbation attack is in bold.

\begin{table}[t]
  \caption{Attack Results on CNN}
  \label{tab:result}
  \footnotesize
  \resizebox{.99\columnwidth}{!}{\begin{tabular}{lccc}
    \toprule
     &IMDB&AMAZON&YELP\\
    \midrule
    CNN classifier accuracy & 97\% & 91\% & 92\% \\
    1\%  perturbation & \textbf{54}\% & 85\% & 81\% \\
    5\%  perturbation & 73\% & 70\% & \textbf{55}\% \\
    10\%  perturbation & 62\% & 58\% & \textbf{18}\% \\
    15\%  perturbation & 61\% & 43\% & \textbf{15}\% \\
  \bottomrule
\end{tabular}}
\end{table}



Unlike the IMDB dataset, the Amazon and Yelp datasets reveal a consistent trend: the exclusion of a greater quantity of words associated with the targeted parts of speech does lead to a decline in classification accuracy. For the Amazon dataset, this decrease is almost linear. However, for the Yelp dataset, there is a sharper decline going from deleting 5\% of the keywords to 10\% of the keywords compared to going from deleting 1\% of the keywords to 5\% of the keywords. 

Overall, outside of the 1\% perturbation outlier on the IMDB dataset, our adversarial example generation framework is consistently the most effective against the Yelp dataset, reducing the accuracy over the baseline by up to 84.53\% at 15\% perturbation. Compared to the other two datasets, over perturbations 5\%, 10\% and 15\%, the accuracy of the CNN classier after the attacks ranges from 15 to 46 percent lower over Yelp than Amazon and IMDB. In contrast, over perturbations 5\%, 10\% and 15\%, our adversarial example generation framework is consistently the weakest against the IMDB dataset, only reducing the baseline accuracy by 44.35\%. 

While the three datasets contain reviews, the reviews cover different domain areas: IMDB contains movie reviews, Yelp contains reviews for businesses, Amazon contains product reviews from a different categories. In addition, the IMDB dataset has fewer reviews compared to the Yelp and Amazon datasets, but the reviews in the IMDB dataset are typically (1) longer and (2) contain more detailed analysis. These dataset attributes combined with our findings indicates that the model's sensitivity to part of speech deletion may be influenced by dataset-specific characteristics and underscores the complex interplay between textual composition and model performance.

\section{Related Work}\label{sec:related}

In this section, we highlight the main areas of work related to adversarial examples.

\textbf{Adversarial examples for text classifications tasks.} 
Adversarial examples were first introduced for text-based systems in~\citeauthor{10.1109/MILCOM.2016.7795300}. Since then, a range of different attacks have been proposed spanning different levels of granularity for what components of the text get modified, ranging from individual characters to whole sentences. 

Character-level attacks like Deep-WordBug are designed to generate adversarial examples in a black-box scenario \cite{gao2018}. Similarly, DISTFLIP uses the HotFlip technique which manipulates individual characters in the input text to create a new text \cite{gil2019}. 

Word-level attacks employ a variety of techniques. \citeauthor{papernot2016} use a computational graph unfolding technique to evaluate the forward derivative, and then employee a fast gradient sign method to guide perturbations. Other works have used modifications like insertion, replacement, and deletion for crafting top-k words \cite{samanta2018}. \citeauthor{sato2018} and \citeauthor{behjati2019} propose adding perturbations in the embedding space and formulate the process as an optimization problem. In addition, several other techniques exist, including optimized word embeddings \cite{ren2019}, attention score calculations \cite{hsieh2019}, iterative replacement \cite{yang2020}, and the use of sememes in word lists \cite{zang2020}. Our work focuses on word-level attacks, since our main investigation surrounds parts of speech.

Sentence-level attacks like the SCPNS generate a paraphrase of the input sentence following the target syntax structure, using LSTM networks with soft attention mechanisms \cite{iyyer2018}. There are also multi-level attacks also exist such as those using FGSM adversarial attack combined with natural language watermarking \cite{liang2018}, the TextBugger framework \cite{li2019}, and reinforcement learning in a black box setting \cite{vijayaraghavan2019}.






\textbf{Adversarial examples for other tasks.} 
Adversarial examples are widely used to test and improve the robustness of models across various tasks, including image classification, natural language processing, and autonomous systems. For example, \citeauthor{lu2019adversarial} targets semantic segmentation and object detection, while \cite{jain2020evading} focuses on bypassing deepfake detection systems. In motion estimation, \citeauthor{usama2019adversarial} explores how adversarial attacks lead to incorrect flow vectors.

In image captioning, \citeauthor{liu2019adversarial} shows how adversarial examples cause models to generate inaccurate captions. Similarly, \citeauthor{agrawal2018adversarial} examines attacks on Visual Question Answering (VQA) models, leading to wrong answers. In natural language processing, \citeauthor{evtimov2018adversarial} highlights vulnerabilities in neural machine translation, and \citeauthor{zhang2019generating} extends attacks to text summarization models.

Additionally, \citeauthor{li2019adversarial} demonstrates how adversarial examples degrade image-to-image translation quality, while \citeauthor{usama2019adversarial2} targets depth estimation in 3D scene understanding. Finally, \citeauthor{fang2020adversarial} and \citeauthor{samadi2019attacks} explore attacks on neural image compression and recruitment ranking systems, respectively. These examples underscore the widespread impact of adversarial attacks across diverse machine learning tasks.
\section{Conclusion}\label{sec:con}
In this paper, we introduce a novel pipeline designed to generate adversarial examples aimed at challenging CNN-based text classification algorithms centered around making manipulations to the text for different Parts-of-Speech. Our approach is evaluated across well-known review datasets -- Yelp, IMDb, and Amazon. Our results highlight the vulnerabilities of CNN models to such adversarial attacks. It is our hope that the methodologies derived from this work will significantly contribute to lowering the barriers for further research into the robustness of CNN-based text classification systems. For our future work, we plan to apply our pipeline to models built on attention and GPT backbones and to compare the outcomes with those of this current study. 

%
%
%
\bibliography{bib}

\end{document}